# Topic Modeling Based Extractive Text Summarization

Kalliath Abdul Rasheed Issam**, Shivam Patel**, Subalalitha C. N.

*Abstract*: *Text summarization is an approach for identifying important information present within text documents. This computational technique aims to generate shorter versions of the source text, by including only the relevant and salient information present within the source text. In this paper, we propose a novel method to summarize a text document by clustering its contents based on latent topics produced using topic modeling techniques and by generating extractive summaries for each of the identified text clusters. All extractive sub-summaries are later combined to generate a summary for any given source document. We utilize the lesser used and challenging WikiHow dataset in our approach to text summarization. This dataset is unlike the commonly used news datasets which are available for text summarization. The well-known news datasets present their most important information in the first few lines of their source texts, which make their summarization a lesser challenging task when compared to summarizing the WikiHow dataset. Contrary to these news datasets, the documents in the WikiHow dataset are written using a generalized approach and have lesser abstractedness and higher compression ratio, thus proposing a greater challenge to generate summaries. A lot of the current state-of-the-art text summarization techniques tend to eliminate important information present in source documents in the favor of brevity. Our proposed technique aims to capture all the varied information present in source documents. Although the dataset proved challenging, after performing extensive tests within our experimental setup, we have discovered that our model produces encouraging ROUGE results and summaries when compared to the other published extractive and abstractive text summarization models.*

## I. INTRODUCTION

In the domain of Natural Language Processing (NLP), text summarization has gained prime importance in recent times. A large amount of text is generated every day in digital form on the internet from mainly news articles, products, service reviews, e-libraries, social media posts, personal and governmental blogs, websites, online tutorials, and e-publications among the few. Though scattered and unprocessed, the text from these sources require computational analysis to gain useful information from it in a quick, efficient and scalable fashion. The novel approaches in the field of text summarization aim to solve this problem.

### A. Text Summarization

Text summarization is a method to extract valuable information from a given text and present it to the user in a simplistic, short and condensed form that retains the relevant content of the source text document to form the summary [1], [2]. High relevance, less redundancy, appropriate compression ratio, and high coverage are the essential factors of a good text summary. The two general classifications of text summarization are Extractive Text Summarization and Abstractive Text Summarization [3], [4], [5]. Section I.B and I.C explain these summarization techniques in brief.

### B. Extractive Text Summarization

In extractive text summarization, as the name suggests, the model extracts salient sentences from the given source document and combines them to form the extractive summary. To select salient sentences, initially, all sentences in the source document are assigned weights, following which, highly ranked sentences based upon their weights are extracted from the source document. These extracted sentences are then combined to generate the text summary [6]. Fig. 1 illustrates the general diagrammatic representation of an extractive text summarization model. The blue lines in the documents shown in the figure depict the original text and the darkish yellow lines in the figure represent the salient sentences selected by the extractive summarizer.

Summaries generated through this approach, generally have low coherence, yet they are highly prevalent in the field of text summarization due to lesser time complexity and greater ease of generation when compared to the summaries generated by the abstractive text summarization approach. The measure of good extractive text summarization is that the generated summary should have proper topic diversity with low redundancy [7], and achieving both of these measures in parallel is highly challenging.

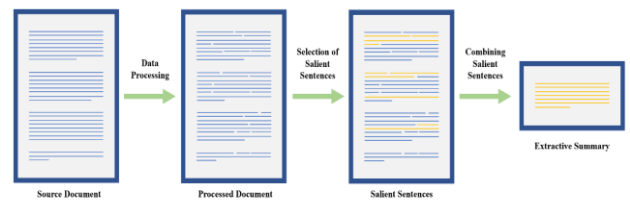

**Fig. 1. General extractive text summarization model**

* Correspondence Author

**Kalliath Abdul Rasheed Issam**, Department of Computer Science and Engineering, SRM Institute of Science and Technology, Kattankulathur, Chennai, India. Email: kalliathissam@gmail.com

**Shivam Patel***, Department of Computer Science and Engineering, SRM Institute of Science and Technology, Kattankulathur, Chennai, India. Email: shivam.patel1606@gmail.com

**Subalalitha C. N.**, Department of Computer Science and Engineering, SRM Institute of Science and Technology, Kattankulathur, Chennai, India. Email: subalaln@srmist.edu.in

** Both authors contributed equally to this research.

## C. Abstractive Text Summarization

In this summarization technique, the summary produced comprises novel sentences which are not extracted from the source document. These sentences convey the main idea of the source document and have been formed by rephrasing the text in it so as to generate summaries that seem more human-written [8]. They are more complex, time-consuming and challenging to generate, and they require extensive analysis and thus are lesser used by people. Though they tend to improve the coherence, readability, and cohesion [8], the summaries generated are plagued by numerous problems such as absurdness, inaccuracy, and repetitiveness [9], [10]. Fig. 2 illustrates the general diagrammatic representation of an abstractive text summarization model. The blue lines in the documents in Fig. 2, represent original text and the green lines depict the novel sentences generated by the abstractive text summarizer.

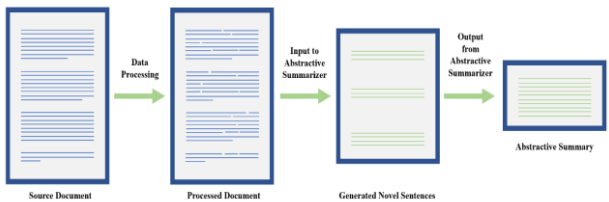

**Fig. 2. General abstractive text summarization model**

## D. Topic Modeling Summarization

In topic modeling text summarization, the main idea is to consider a document comprising of various *topics* [11]. Using topic modeling techniques such as LDA, the *topics* in the source document are identified. These *topics* are then used to generate text clusters. The clusters include salient sentences from the source document. Each cluster would be linked to the relevant identified *topics*. Typically, this summarization approach designates the sentences in a source document to various identified *topics*, to increase coverage and thus help in the summarization of the document [12]. The prime advantage of this approach is to considerably enhance the degree of topic selection from the source document, which will thus generate better summaries.

## II. LITERATURE REVIEW

### A. Single Document Summarization

During the initial research in text summarization, summarizing single documents was the primary focus of scholars. The technique of identifying salient information from a single text document, thus compressing and summarizing it, is called single-document summarization. A lot of the state-of-the-art models for single document summarization limit their functioning to short length text and hence do not produce encouraging results when applied over longer length text.

Romain P. et al. [13] propose a neural network model having intra-attention to generate abstractive summaries for input documents. They have used reinforcement learning and a supervised algorithm to predict words for creating novel phrases for their abstractive summary. With this combination, they claim to generate more readable summaries. However, they have trained and evaluated their model on different datasets than ours, using the standard CNN/Daily Mail and New York Times datasets. We have used the WikiHow dataset, which is very less explored in terms of text summarization research.

The WikiHow dataset presents higher abstractedness than the CNN/Daily Mail dataset, which makes the summarization of its documents more challenging as the model needs to be extra creative in generating unique summaries [14]. Abstractedness is a measure of the novel n-grams present in the reference summary given with the dataset which (novel n-grams) are absent in the source document. A graph comparing the level of abstractedness between WikiHow and CNN/Daily Mail dataset is shown in Fig. 3. This graph is taken from [14].

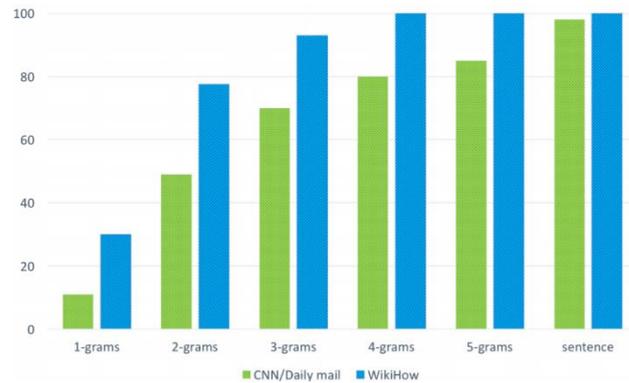

**Fig. 3. Uniqueness of n-grams in CNN/Daily Mail and WikiHow datasets**

Moreover, the WikiHow dataset has a higher compression ratio than the CNN/Daily Mail dataset. The compression ratio is calculated by dividing the average length of sentences in a dataset with the average length of reference summaries given with the dataset. Table- I shows the compression ratio statistics for both these datasets. These statistics are referred from [14]. Having a higher compression ratio makes summarizing the WikiHow documents a more difficult task when compared to summarizing the documents of the CNN/Daily Mail dataset. This is because, with a higher compression ratio, models for summarizing WikiHow documents would need to obtain more prominent semantics and abstraction. Further discussion on how our dataset is different from the other standard text summarization datasets and why generating text summaries on our dataset is more challenging than for other standard datasets is argued in Section IV.A.

**Table- I: Compression ratio of WikiHow and CNN/Daily Mail datasets. The represented article and summary lengths are the averages over all sentences.**

|  | WikiHow | CNN/Daily Mail |
|---|---|---|
| **Article Sentence Length** | 100.68 | 118.73 |
| **Summary Sentence Length** | 42.27 | 82.63 |
| **Compression Ratio** | 2.38 | 1.44 |

An extractive summarization model on a single text document as a tree induction scenario is formulated by Yang L. et al. [15]. Here, the root nodes in the tree represent summary sentences. Each root node has attached sub-trees that contain phrases or words to elaborate on the root node summary sentence. Their model is iterative in nature which helps to generate better summaries in each iteration, but again, they limit their evaluation to the standard datasets used by Romain P. et al. in [13], i.e. CNN/Daily Mail and New York Times datasets.

A single document extractive summarization is conceptualized by Sanchit A. et al. [16]. It uses sentence embeddings and K-Means clustering. Using K-Means clustering, they form clusters of text for each document, such that the intra-cluster similarity is higher than the inter-cluster similarity. Following this, the most informative and important sentence from each cluster is chosen to generate an extractive summary for a given source document. Though this approach is effective, their model varies from ours. We have used a different approach to generate clusters and have trained our model on a more challenging dataset.

Naveen S. et al. [17] present an interesting model for generating single document extractive summaries by considering the text summarization problem to be one of the binary optimization problems, which uses MOBDE (Multi-Objective Binary Differential Evolution) to generate text summaries. They have used various statistical features such as coverage, sentence length, the position of a sentence in the document, cohesion, similarity between title and sentence, and readability score, to evaluate summary sentences. Moreover, when evaluating their model on DUC2001 and DUC2002 standard text summarization datasets, they obtain comparative results to other state-of-the-art techniques.

**B. Multi-Document Summarization**

The text summarization approach which generates a summary for a topic after considering multiple documents about the topic is called multi-document summarization. In this summarization technique, relevant sentences for the topic are either picked up (for extractive text summarization) or are rephrased (for abstractive text summarization) after performing multi-document analysis. The sentences are then combined to form a representative summary for the set of documents relevant to a topic.

In [18], Jian-Ping M. et al. developed a multi-document summarization framework to generate extractive summaries. They introduced two new features to calculate the scores of sentences in the summaries. The first being 'Exemplar', a feature to balance the relevance and coverage in their summaries and the second being 'Position', to measure the relative position of each sentence in a document. However, along with evaluating their model on the standard DUC datasets (DUC2004, DUC2005, DUC2006), they have used the ROUGE-2 and ROUGE-SU4 scores to compare their summaries with those generated by the other state-of-the-art techniques. Moreover, this comparison shows a very marginal increase in the performance of their approach.

An evolutionary algorithm is proposed by Rasim M. A. et al. [19] to perform summarization on multiple documents. They claim their model to have a superior correlation between sentences with a low rate of redundancy. Nevertheless, their model requires a large computational overhead and is experimented over the standard DUC2002 and DUC2004 datasets to get viable results. Aiming to have good topic coverage, this approach cannot be directly applied to WikiHow dataset for generating topic-based extractive summaries.

A multi-document summarization technique to summarize documents containing events that have occurred in the past at different times was introduced in [20] by Giang T. et al. This simple timeline summarization approach aims to summarize prolonged events such as economic crises and war. Though it achieves good performance on timeline data, this method will not present equivalent results on the WikiHow dataset. This is because the task of forming text clusters on a well-defined timeline dataset and generating summaries is different and less challenging than performing a similar task on a dataset with a high diversity of topics and novel writing style.

Libin Y. et al. [21] present the technique of multi-document summarization using topic ranking and by generating topic clusters. Their model is successful is reducing redundancy in the generated summaries and it also produces summaries that are of high quality. However, since it lacks topic diversity, it presents challenges to generate WikiHow summaries of high coverage.

**C. Topic Modeling Summarization**

The objective of this summarization approach is to form clusters of text from the content present in a source document and summarize the clusters to generate the document summary. In order to achieve good results in this summarization approach, the source document must have a high topic-diversity. Various topic modeling techniques exist that help to enhance the topic selection quality.

A novel topic augmented abstractive summarization method that identifies the topics in the source document using LDA has been introduced by Melissa A. et al. [22]. Moreover, they have used CNN/Daily Mail and WikiHow datasets for evaluation. They have performed five different implementations, i.e., (i) Pointer-Generator, (ii) TAG, (iii) Pointer-Generator + Coverage, (iv) TAG + Coverage and (v) Lead-3, on the WikiHow dataset. Our model outperforms (i), (ii), (v) and gets an equivalent ROUGE for (iii).

Liu N. et al. [23] propose an algorithm to summarize multi-document text using LDA topic modeling. Among the topics identified by LDA, their model selects essential topics based on weight criteria. They have also used statistical features such as sentence length, frequency of terms and sentence position to train their models. Evaluating their model on the standard DUC2002 corpus, they claim to achieve stronger results than other state-of-the-art algorithms.

An unsupervised approach is presented by Lu W. et al. [24], to summarize spoken meetings using topic modeling. Their method exceeds the summarization results of existing models and eliminates redundancies. Their token-level summarization framework uses utterance-level topic modeling and is claimed to perform better than other document-level models.

Zongda W. et al. [25] introduces a robust framework to summarize novel documents. This well explained research work uses topic modeling to create text clusters from which salient candidate sentences are selected based upon importance scores, to generate text summaries. Trained directly on 63 narrative novels, this model is suitable mainly for summarizing long-form novel datasets and is not appropriate for summarizing documents present in the WikiHow dataset. Post generating their initial summaries, they also smooth their summaries to make it more readable.

An approach to extractive text summarization using binary classification modeling and topic-based sentence extraction using LDA is proposed in the research work by Nikolaos G. et al. [26]. Their evaluation of the MultiLing 2015 MSS dataset proves that topic modeling helps generate better extractive text summaries. However, they use supervised learning to perform binary classification on sentences, where one class holds sentences to be included in the summary and the other class includes the sentences that should not be present in the summary. Contrary to it, we propose a completely unsupervised approach to generated extractive summaries.

### III. SYSTEM ARCHITECTURE

#### A. System Design

The system design of our proposed model for executing topic modeling based extractive text summarization is described below in Fig. 4. We utilize the WikiHow dataset [14] for this project. Initially, the dataset is processed to generate separate folders for the articles and their reference summaries. The articles undergo data cleaning, wherein they are tokenized, following which, the unnecessary characters present within them are stripped off, and finally converted into a list of token sequences.

The list of token sequences is used to generate the dictionary and corpus for the dataset. Following this, our Latent Dirichlet Allocation (LDA) [11] model takes the generated dictionary and corpus as its input. The LDA model produces a list of abstract topics that classifies the tokens in the dataset. The trained LDA model classifies sentence tokens based on their most characteristic topic. This topic-based distribution of sentences forms topic-level clusters for each document present in the dataset.

The per-document topic-level clusters are used to generate subdocument summaries using the TextRank algorithm [27]. The subdocument summaries are combined to generate the document summary for each document. We use the ROUGE [28] metric to compare the generated summaries with the reference summaries giving in the WikiHow dataset.

#### B. Sequence Diagram

As seen in Fig. 5, our model consists of five modules, namely (i) Data Cleaning Module, (ii) Topic Modeling Module, (iii) Topic Clustering Module, (iv) Document Summarization Module, and (v) Evaluation Module.

The Data Cleaning Module, being the first module, pre-processes the articles present in the dataset. Pre-processing of articles includes tokenization, stop word removal, bigram generation, trigram generation, and lemmatization. The Topic Modeling Module generates the dictionary and corpus that act as input to the LDA model, which produces the list of abstract topics representative of the whole dataset. The Topic Clustering Module is responsible for the generation of per-document topic-level clusters for all documents in the dataset. Following this, the Document Summarization Module summarizes each subdocument separately using the TextRank algorithm, an extractive summarization technique. The summaries generated for each subdocument are combined to form a complete extractive summary for the given document. Finally, the Evaluation Module compares summaries generated by our model with the reference summaries given in the WikiHow dataset. The evaluation metric used by us is ROUGE which measures our performance on summarizing the articles present in the WikiHow dataset.

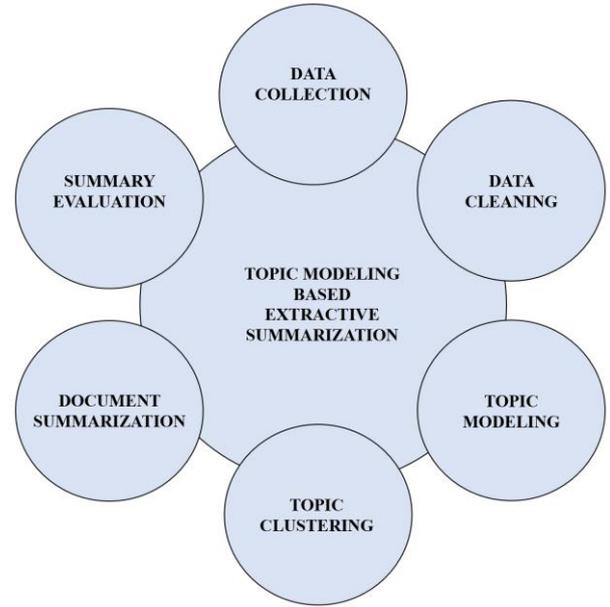

**Fig. 4. Proposed system design for topic modeling based extractive text summarization**

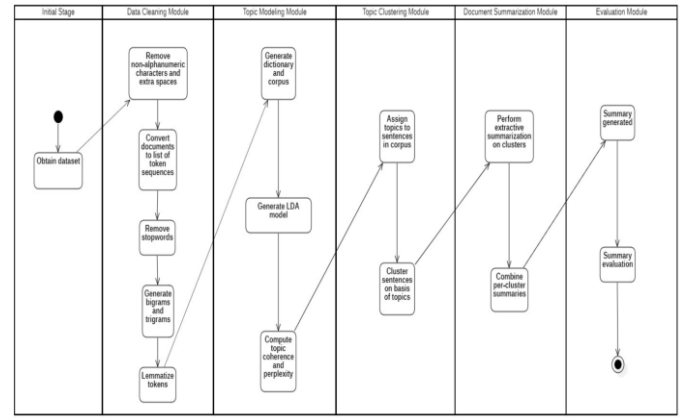

**Fig. 5. Sequence diagram for the proposed methodology**

## IV. DATASET

### A. Dataset Explanation

As mentioned earlier, we have used the WikiHow dataset in our approach to text summarization. This dataset contains articles found on the WikiHow website [29], which is an online community composed of how-to guides for learning to do anything. The WikiHow dataset is a diverse dataset consisting of articles on a wide variety of topics which are written by a vast number of distinct authors. Furthermore, ordinary people are the authors of these articles, and the thus language used in these articles is neither formal nor complex and can be easily understood.

These articles consist of steps to complete a given task. The task can be solved in a single approach or it can be achieved in multiple ways. Thus, the articles sometimes contain a single method to solve a given task or it can describe multiple approaches to solving the task. The statistics of the WikiHow dataset are presented in Table- II. Moreover, the articles comprise of multiple step-like subheadings followed by a paragraph explaining each step. The dataset has used paragraphs and subheadings respectively to generate the document and summary for each article.

**Table- II: WikiHow dataset characteristics**

| Dataset Size | 230,843 |
|---|---|
| Average Article Length | 579.8 |
| Average Summary Length | 62.1 |
| Vocabulary Size | 556,461 |

The WikiHow dataset is novel in comparison to other text summarization datasets available online. Unlike most text summarization datasets, the WikiHow dataset is not a news dataset. Thus, there is a marked difference in the generation of summaries from the WikiHow dataset and other news datasets. News datasets follow a specific format for the distribution of information. News articles present most of the essential information in the first few sentences. Therefore, it is possible to obtain relatively good summaries just by taking the first three sentences of the article. Rather than presenting the essential information at the beginning, the WikiHow articles present information uniformly throughout. Therefore, we need to develop better summarization models that are capable of generating summaries universally, irrespective of domain.

Moreover, the WikiHow dataset is appropriate for generating abstractive summaries. Although its summaries are semantically similar to its documents, they do not share the exact words. Level-of-abstractedness and compression ratio are the two metrics introduced to highlight the abstractive nature of the WikiHow dataset. The level-of-abstractedness metric focuses on the number of shared n-grams present in the reference summaries and the original articles. The compression ratio metric refers to the ratio of the average article length to the average reference summary length. The low-valued level-of-abstractedness and high-valued compression ratio scores demonstrate clearly that the WikiHow dataset furnishes new challenges to researchers in the field of text summarization. A sample document present on the WikiHow website [30] is shown in Fig. 6 and its corresponding processed document as present in the WikiHow dataset is shown in Fig. 7.

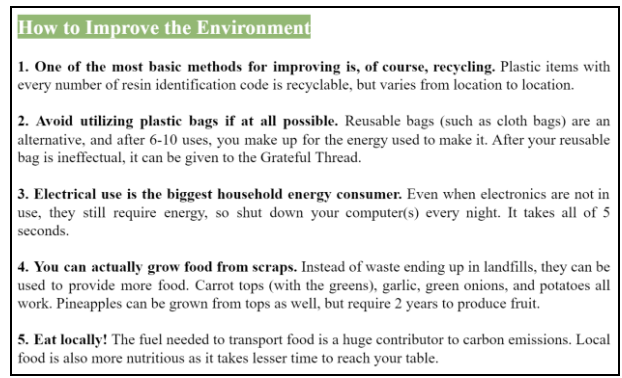

Fig. 6. An example of a WikiHow document

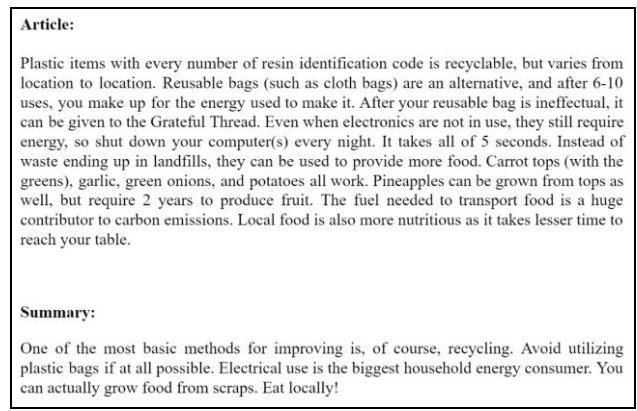

Fig. 7. Processed article and summary for the WikiHow document shown in Fig. 6

## V. PROPOSED METHODOLOGY

### A. Data Cleaning

The architecture diagram of the proposed work of this paper is shown in Fig. 8. This section describes the pre-processing steps applied to the WikiHow dataset to capture high-quality topics from it. It starts with removing the stop words and the unnecessary characters from the source documents in the dataset, followed by tokenizing, generation of bigrams and trigrams and eventually lemmatizing the tokens. Experimental results have shown that summaries generated after cleaning the source data are much better than without executing this step.

1) **Removal of stop words and unnecessary characters:** In this step of data cleaning, all words that do not contribute to the improvement of the topic modeling process are removed. Examples of such words are connectives and high-frequency words present in the dataset. Unnecessary spaces and other non-alphanumeric characters are also removed to improve topic modeling results.
2) **Tokenization:** Each document present in the dataset is tokenized to reduce the complexity of words in it. The tokenization process involves the conversion of each word to its lowercase form and removal and replacement of all accented characters.
3) **Generation of bigrams and trigrams:** Bigrams and trigrams are generated from the tokenized dataset so that the phrases which occur frequently in the dataset can be considered as a single unit. The generation of bigrams and trigrams depends on their frequency of occurrence in the document. Once their occurrence frequency exceeds a specified minimum value and while it remains within a particular threshold parameter, these bigrams and trigrams are generated.
4) **Lemmatization of tokens:** Lemmatization is the process of converting each token to its root form. For our model, we have included only those tokens whose part-of-speech tag belonged to a specified group of nouns, verbs, adjectives or, adverbs. For each document, all qualifying tokens were converted to their lemmatized forms and combined to create a list of token sequences for the entire dataset.

B. Topic Modeling

1) **Generation of dictionary and corpus:** Processed and cleaned data is then used to produce the required dictionary and corpus. This is an important step in creating topic model distributions for our dataset. The dictionary maps the tokens of the dataset to their respective token IDs. And the corpus refers to a bag-of-words representation of the tokens available in the dataset, i.e., it contains tuples consisting of a token ID and its respective token count for every token present in the dataset.
2) **Generation of Latent Dirichlet Allocation model:** Latent Dirichlet Allocation (LDA) [11] is a generative probabilistic model used to determine the abstract topics that are present in the dataset. A Dirichlet distribution refers to a distribution of distributions. Thus, LDA is a statistical technique that maps the distribution of documents and abstract topics to the distribution of words and abstract topics. Each topic is latent, i.e., hidden and is represented by a distribution of words present in the dataset. Hence, each document in the dataset is defined as the distribution of such topics according to LDA. Another version of Latent Dirichlet Allocation is known as MALLET LDA. MALLET LDA utilizes Gibbs sampling and efficient document-topic hyperparameter optimization to generate topics from the dataset. In our model, we have implemented Gensim LDA [31] as well as MALLET LDA [32] with different numbers of topics.
3) **Computation of perplexity and coherence:** Perplexity is a measure of how perplexed or surprised a text summarization model is upon receiving previously unseen data from a test set. It is defined as the normalized log-likelihood of a held-out test set. Although the perplexity metric presents little information on its own, it is a useful metric for the comparison of different models. The lower the perplexity of a model, the better will be the topic quality. The coherence of a model measures the semantic similarity of the top words for each topic. Thus, higher the coherence value of a text summarization model, the better its topic quality.

C. Topic Clustering

To generate topic clusters, we tokenize each document into sentences. Each document of the dataset is converted to a sentence group, i.e., a list of tuples of sentence number along with its respective sentence. With the help of the trained LDA model, every sentence in each sentence group gets assigned with their most representative topic ID. This information is also stored onto a data frame for faster processing in future steps.

By using the previously-constructed data frame, each document gets transformed into clusters of sentences for every topic. This results in the generation of a list of topic clusters. The topic clusters then map each topic to a collection of sentences for each document. For better processing of data in future steps, a list of topic IDs representative of a given document is generated for each document in the dataset.

D. Document Summarization

The generated topic clusters for a document tend to divide the document into multiple sub-documents. These sub-documents are then summarized using TextRank, a robust extractive text summarization technique.

The TextRank algorithm is an adaptation of the PageRank algorithm. It differs from the PageRank algorithm, as unlike PageRank which is used to rank web pages, TextRank is used for ranking sentences. The algorithm combines the text in all documents and tokenizes them on a sentence-level basis. Following this, it converts the sentences into vectorized forms by using word embeddings such as GloVe word embeddings. The algorithm then uses the vectors to construct a similarity matrix using cosine similarity. Finally, it develops a ranked graph of sentences using the similarity matrix and generates the summary by choosing the top N sentences according to specified parameters.

The summaries created from each sub-document are combined to generate the summary for a given text document. However, when the sub-documents are of a shorter length than the specified threshold, our model includes the entire sub-document as part of the document summary. The algorithm of our proposed work is shown below Fig. 8.

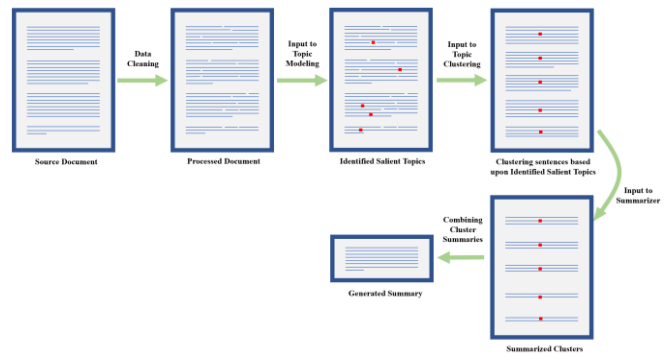

**Fig. 8. Architecture for the proposed methodology**

**Algorithm:** Topic Modeling Based Extractive Text Summarization

---

**Input:** WikiHow articles

**Output:** Summaries for WikiHow articles

**Begin**
1. **SET** cleaned_dataset <= []
2. **FOR EACH** document ∈ WIKIHOW_ARTICLES
3.    **SET** cleaned_data <= REMOVE-UNNECESSARY-CHARACTERS (document)
4.    **UPDATE** cleaned_data <= TOKENIZE-AND-REMOVE-STOPWORDS (cleaned_data)
5.    **UPDATE** cleaned_data <= GENERATE-BIGRAMS-AND-TRIGRAMS (cleaned_data)
6.    **UPDATE** cleaned_data <= LEMMATIZE-TOKENS (cleaned_data)
7.    **ADD** cleaned_data TO cleaned_dataset
8. **ENDFOR**

9. **SET** dataset_dictionary <= GENERATE-DICTIONARY (cleaned_dataset)
10. **SET** dataset_corpus <= GENERATE-CORPUS (cleaned_dataset)
11. **SET** lda_model <= GENERATE-LDA-MODEL (dataset_dictionary, dataset_corpus)
12. **SET** dataset_sentence_groups <= []
13. **SET** dataset_sentence_distributions <= []

14. **FOR EACH** document ∈ WIKIHOW_ARTICLES
15.    **SET** doc_sentences <= TOKENIZE-SENTENCES (document)
16.    **ADD** doc_sentences TO dataset_sentence_groups
17.    **SET** document_sentence_distribution <= []
18.    **FOR EACH** sentence ∈ doc_sentences
19.      **SET** dominant_topic_id <= GENERATE-DOMINANT-TOPIC (lda_model, sentence)
20.      **ADD** (sentence, dominant_topic_id) TO document_sentence_distribution
21.    **ENDFOR**
22.    **ADD** document_sentence_distribution TO dataset_sentence_distributions
23. **ENDFOR**

24. **SET** dataset_topic_clusters <= []
25. **FOR EACH** document_sentence_distribution ∈ dataset_sentence_distributions
26.    **SET** document_topic_cluster <= GENERATE-TOPIC-CLUSTERS (document_sentence_distribution)
27.    **ADD** document_topic_cluster TO dataset_topic_clusters
28. **ENDFOR**

29. **SET** generated_summaries <= []
30. **FOR EACH** document_topic_cluster ∈ dataset_topic_clusters
31.    **SET** document_summary <= []
32.    **FOR EACH** (topic_id, topic_content) ∈ document_topic_cluster
33.      **SET** topic_summary <= TEXTRANK-SUMMARIZE (topic_content)
34.      **UPDATE** document_summary <= document_summary + topic_summary
35.    **ENDFOR**
36.    **ADD** document_summary TO generated_summaries
37. **ENDFOR**
**End**

## VI. PERFORMANCE METRICS

### A. ROUGE

The ROUGE (Recall-Oriented Understudy for Gisting Evaluation) [28] is the most popular set of metrics used for evaluating summaries produced by machines. It compares the base summaries (reference summaries or true summaries) present in the source dataset with the summaries generated automatically by machine models. The base summaries are generally written by humans.

The ROUGE metric set presents five different evaluation metrics. ROUGE-N (ROUGE-1, ROUGE-2 and so on) measures the overlap between n-grams from the base summary with those of the automatically generated summary. ROUGE-L (R-L) is a metric that measures the longest co-occurring n-grams between the reference and machine summary. Furthermore, ROUGE-W (R-W) evaluates the weighted longest common subsequence between the summaries presented for evaluation. The ROUGE metric set also contains ROUGE-S (R-S) which stands for skip-bigram based co-occurrence for the tokens. Finally, the fifth ROUGE metric is ROUGE-SU (R-SU) which measures skip-bigram along with unigram co-occurrences in summary evaluation. The general ROUGE equation is presented below.

$$ROUGE - N = \frac{\sum_{S \in \{Reference\ Summaries\}} \sum_{gram_n \in S} Count_{match}(gram_n)}{\sum_{S \in \{Reference\ Summaries\}} \sum_{gram_n \in S} Count(gram_n)}$$

We have evaluated the WikiHow summaries generated by our model using ROUGE-1 (R-1), ROUGE-2 (R-2), ROUGE-3 (R-3), ROUGE-4 (R-4), ROUGE-L and ROUGE-W, and have considered the precision, recall and F-1 scores values for each of these evaluation metrics. A recall of 35% on the ROUGE-N metric states that the generated summary contains 35% of n-grams present in base summary. A precision score of 47% on the ROUGE-N metric claims that 47% of the n-grams present in the generated summary are also found in the base summary. The F-1 score is the harmonic mean of the calculated recall and precision values. However, it is difficult to elucidate the literal meaning of x% F-1 score on a ROUGE-N metric.

### B. Discussion on Performance Metrics

Though ROUGE scores are the most used metric in evaluating text summarization research, they do not present the best way to measure the quality of machine-generated summaries. Generating a summary for any given text document is very subjective. The summaries generated for any given document by different people will be dissimilar to each other as they may use completely different words, phrases or sentence structures to bring out the same meaning. In such cases, the summary with a greater overlap of its n-grams with those in the reference summary will have higher ROUGE scores when compared to the summary with lesser overlapping n-grams.

Thus, ROUGE score does not tend to capture the actual summarization quality as there are multiple valid ways to summarize a text. In summarization, it is important to identify the salient ideas in the source text and phrase it accurately to generate the summary. But ROUGE does not tolerate rephrasing of sentences present in the reference summary and assigns a higher score for summaries that have phrases identical to those present in the reference summaries [10]. Thus, a better text summarization evaluation metric needs to be developed which considers the limitations of ROUGE.

## VII. RESULTS & DISCUSSIONS

### A. ROUGE Score Results

This section represents the evaluation of the WikiHow summaries generated by our model with the reference summaries given in the WikiHow dataset.

We have compared our summarization results on the WikiHow dataset with other famous state-of-the-art methods for text summarization on this dataset. The results of our experiments are described in Table- III. It is encouraging to state that the proposed methodology of this paper has surpassed most other methods of summarization and has marginal differences in ROUGE scores from the few better performing models. However, as described in the previous section, summaries generated by any model are always subjective and hence cannot be best judged by the famous above-stated metric. Our model claims to generate summaries that include most of the important topics and salient information contained in the source document and present a considerable and convincing summary to the user.

**Table- III: Comparison of ROUGE metric performance of various text summarization models on WikiHow dataset**

| Model | WikiHow Dataset | | |
|---|---|---|---|
| | ROUGE-1 | ROUGE-2 | ROUGE-L |
| Seq-to-seq with attention [14] | 22.04 | 6.27 | 20.87 |
| Lead-3 [22] | 26.00 | 7.24 | 24.25 |
| Pointer-Generator [22] | 26.02 | 7.92 | 24.59 |
| Topic Augmented Generator [22] | 26.18 | 8.18 | 25.25 |
| Pointer-Generator + Coverage [22] | 27.08 | 8.49 | 26.25 |
| **Topic Modeling Based Extractive Text Summarization Model** (this paper) | **27.08** | **6.89** | **25.43** |
| TextRank [14] | 27.53 | 7.4 | 20.00 |
| Topic Augmented Generator + Coverage [22] | **28.36** | **9.05** | **27.48** |

### B. Topic Modeling Based Extractive Summarization Output on Sample WikiHow Document

Through this section, we aim to better explain the functioning of our proposed work by performing text summarization on a sample WikiHow document. The document is titled 'How to Comb Long Hair' and is taken from [33]. The entire document as taken from [33] is shown in Fig. 9, and the processed document as present in the WikiHow dataset is displayed in Fig. 10.

After processing the text as shown in Fig. 10 through our Data Cleaning Module, the cleaned text that remains is presented in Fig. 11. On this cleaned text, we run our Topic Modeling Module to identify the multiple latent topics present in the source document.

Post this, our Topic Clustering Module generates text clusters equivalent in quantity to the number of identified topics. The clusters are generated around the discovered topics and each cluster contains essential text from the source document which is relevant to its respective identified topic. The text clusters created by our model for the WikiHow document shown in Fig. 9 are displayed in Fig. 12.

Finally, our model generates summaries for each of the identified text clusters which are eventually combined to form the 'Generated Summary' for the input document taken into consideration for summarization. Fig. 13 shows the 'Given Summary' and 'Generated Summary' for the sample WikiHow document being considered. The 'Given Summary' is the summary present in the official WikiHow dataset and the summary created by our proposed approach is shown under 'Generated Summary'.

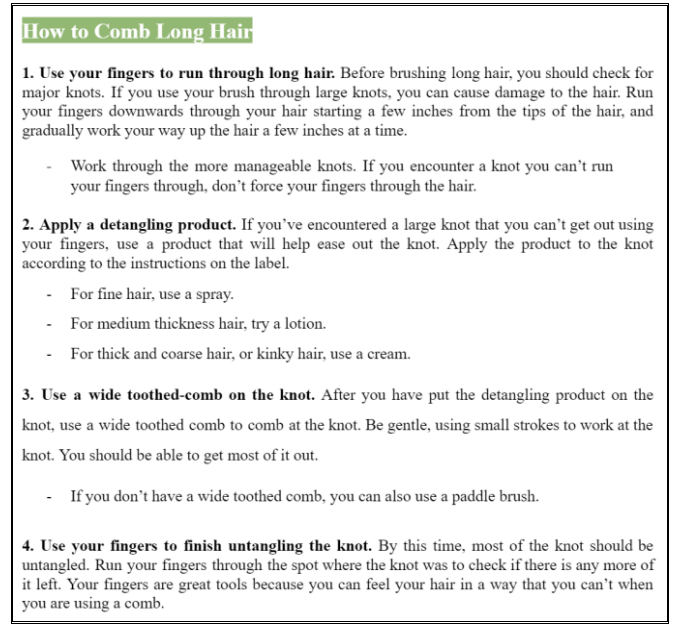

**Fig. 9. Sample document from WikiHow website [33]**

It is seen that the 'Given Summary' has used considerably short abstractive sentences to summarize the source document. Conversely, as expected, the summary generated by our model, being an extractive approach, is of a bit longer length than the one given in the WikiHow dataset. But it is interesting to know that all the salient topics in the input document as well as in its official summary, have been included in the summary generated by our model.

**Article:**

Before brushing long hair, you should check for major knots. If you use your brush through large knots, you can cause damage to the hair. Run your fingers downwards through your hair starting a few inches from the tips of the hair, and gradually work your way up the hair a few inches at a time. Work through the more manageable knots. If you encounter a knot you can't run your fingers through, don't force your fingers through the hair. If you've encountered a large knot that you can't get out using your fingers, use a product that will help ease out the knot. Apply the product to the knot according to the instructions on the label. For fine hair, use a spray. For medium thickness hair, try a lotion. For thick and coarse hair, or kinky hair, use a cream. After you have put the detangling product on the knot, use a wide toothed comb to comb at the knot. Be gentle, using small strokes to work at the knot. You should be able to get most of it out. If you don't have a wide toothed comb, you can also use a paddle brush. By this time, most of the knot should be untangled. Run your fingers through the spot where the knot was to check if there is any more of it left. Your fingers are great tools because you can feel your hair in a way that you can't when you are using a comb.

**Fig. 10.** Document in Fig.9 as present in the WikiHow dataset

**Cleaned Data:**

brush long hair check major knot brush large knot damage hair run finger hair start inch tip hair gradually work hair inch work manageable knot encounter knot run finger force finger hair encounter large knot finger product ease knot apply product knot accord instruction label fine hair spray medium thickness hair lotion thick coarse hair kinky hair cream put detangle product knot comb knot gentle small stroke work knot able wide_toothed_comb paddle brush knot untangled run finger spot knot check finger great tool hair comb

**Fig. 11.** Document in Fig.9 after processing through the Data Cleaning Module

**Generated Clustered Sentences per Topic:**

{0: ['Before brushing long hair, you should check for major knots.'],
1: [],
2: ['If you use your brush through large knots, you can cause damage to the hair.',
'Apply the product to the knot according to the instructions on the label.',
'For fine hair, use a spray.',
'For medium thickness hair, try a lotion.',
'For thick and coarse hair, or kinky hair, use a cream.',
'Your fingers are great tools because you can feel your hair in a way that you can't when you are using a comb.'],
3: [], 4: [], 5: [],
6: ['Run your fingers downwards through your hair starting a few inches from the tips of the hair, and gradually work your way up the hair a few inches at a time.',
'Work through the more manageable knots.',
'If you encounter a knot you can't run your fingers through, don't force your fingers through the hair.',
';\n'
'If you've encountered a large knot that you can't get out using your fingers, use a product that will help ease out the knot.',
'After you have put the detangling product on the knot, use a wide toothed comb to comb at the knot.',
'Be gentle, using small strokes to work at the knot.',
'If you don't have a wide toothed comb, you can also use a paddle brush.',
'By this time, most of the knot should be untangled.',
'Run your fingers through the spot where the knot was to check if there is any more of it left.'],
7: [], 8: [],
9: ['You should be able to get most of it out.'],
10: [], 11: [], 12: [], 13: [], 14: [], 15: [], 16: [], 17: [], 18: [], 19: []}

**Fig. 12.** Text clusters generated by our model for the article in Fig. 9

**Given Summary:**

Use your fingers to run through long hair. Apply a detangling product. Use a wide toothed-comb on the knot. Use your fingers to finish untangling the knot.

**Generated Summary:**

If you use your brush through large knots, you can cause damage to the hair. For fine hair, use a spray. If you encounter a knot you can't run your fingers through, don't force your fingers through the hair. If you've encountered a large knot that you can't get out using your fingers, use a product that will help ease out the knot. After you have put the detangling product on the knot, use a wide toothed comb to comb at the knot. You should be able to get most of it out.

**Fig. 13.** Reference summary given for article present in Fig. 9 (under the heading 'Given Summary') and its summary generated by our model (under the heading 'Generated Summary')

## VIII. CONCLUSION & FUTURE ENHANCEMENTS

We have proposed a novel extractive text summarization method that builds upon the existing summarization techniques. The existing techniques usually favor brevity instead of incorporating all the essential information present within the source text documents. We used topic modeling to identify salient topics present in a document to be summarized and have generated text clusters around those topics. The summaries generated for each of these text clusters are combined to form the final summary of the input document. Our dataset consists of very short abstractive summaries, thus proving itself to be extremely challenging to produce high ROUGE scores. But it is encouraging to realize that our model has performed comparatively well with respect to the other published extractive and abstractive summarization models.

The ultimate aim of our work is to achieve human-level summarization. In order to accomplish this, it is imperative to create better abstractive models for text summarization. We also plan to improve the ROUGE scores of our model by iterating through a larger number of different hyperparameters. Apart from that, we aim to research to develop a better metric than ROUGE, or an improvement on ROUGE to evaluate the quality of summaries generated by different models. We agree that it is a challenging task, as summaries of a document are always subjective, and may have completely different words presenting the same idea to the reader. But this future work, if implemented, would be a major contribution to this emerging field of text summarization.

**REFERENCES**

1. S. Soumya, Geethu S. Kumar, Rasia Naseem, and Saumya Mohan, 2011, "Automatic Text Summarization", In: Das V.V., Thankachan N. (eds), Computational Intelligence and Information Technology. CIIT 2011, Communications in Computer and Information Science (CCIS), Springer, Berlin, Heidelberg, Volume 250, pp. 787-789.
2. Rasim M. Alguliyev, Ramiz M. Aliguliyev, Nijat R. Isazade, Asad Abdi, and Norisma Idris, 2019, "COSUM: Text summarization based on clustering and optimization", Expert Systems: The Journal of Knowledge Engineering, Volume 36, Issue 1.
3. Hongyan Jing, 2000, "Sentence Reduction for Automatic Text Summarization", Sixth Applied Natural Language Processing Conference. Association for Computational Linguistics, pp. 310–315.
4. Kevin Knight, and Daniel Marcu, 2002, "Summarization beyond sentence extraction: A probabilistic approach to sentence compression", Artificial Intelligence, Volume 139, Issue 1, Pages 91-107.

5. Mani I., 2001, "Automatic summarization", John Benjamin's Publishing Company, Amsterdam/Philadelphia.
6. J.N. Madhuri, and R. Ganesh Kumar, 2019, "Extractive Text Summarization Using Sentence Ranking", 2019 International Conference on Data Science and Communication (IconDSC), IEEE, pp. 1-3.
7. Mahak Gambhir, and Vishal Gupta, 2017, "Recent automatic text summarization techniques: a survey", Artificial Intelligence Review, Volume 47, Issue 1, pp 1–66.
8. Panagiotis Kouris, Georgios Alexandridis, and Andreas Stafylopatis, 2019, "Abstractive Text Summarization Based on Deep Learning and Semantic Content Generalization", Proceedings of the 57th Annual Meeting of the Association for Computational Linguistics, Association for Computational Linguistics, pp. 5082–5092.
9. Soumye Singhal, and Arnab Bhattacharya, 2015, "Abstractive Text Summarization", pp. 1-11.
10. Abigail See, Peter J. Liu, and Christoper D. Manning, 2017, "Get To The Point: Summarization with Pointer-Generator Networks", Proceedings of the 55th Annual Meeting of the Association for Computational Linguistics (Volume 1: Long Papers), pp. 1073-1083.
11. David M. Blei, Andrew Y. Ng, and Michael I. Jordan, 2003, "Latent Dirichlet Allocation", Journal of Machine Learning Research, Volume 3, pp. 993-1022.
12. Zhou Tong, and Haiyi Zhang, 2016, "A Text Mining Research Based on LDA Topic Modeling", The Sixth International Conference on Computer Science, Engineering and Information Technology, Volume 6, pp. 201-210.
13. Romain Paulus, Caiming Xiong, and Richard Socher, 2018, "A Deep Reinforced Model for Abstractive Summarization", International Conference on Learning Representations.
14. Mahnaz Koupaee, and William Wang, Oct 2018, "WikiHow: A Large-Scale Text Summarization Dataset", ArXiv.
15. Yang Liu, Ivan Titov, and Mirella Lapata, 2019, "Single Document Summarization as Tree Induction", Association for Computational Linguistics, Proceedings of the 2019 Conference of the North American Chapter of the Association for Computational Linguistics: Human Language Technologies, Volume 1 (Long and Short Papers), pp. 1745-1755.
16. Sanchit Agarwal, Nikhil Kumar Singh, and Priyanka Meel, 2018, "Single-Document Summarization Using Sentence Embeddings and K-Means Clustering,", 2018 International Conference on Advances in Computing, Communication Control and Networking (ICACCCN), pp. 162-165.
17. Naveen Saini, Sriparna Saha, Dhiraj Chakraborty, and Pushpak Bhattacharyya, 2019, "Extractive single document summarization using binary differential evolution: Optimization of different sentence quality measures", PLOS ONE 14(11): e0223477.
18. Jian-Ping Mei, and Lihui Chen, 2012, "SumCR: A new subtopic-based extractive approach for text summarization", Knowledge and Information Systems, Volume 31, Issue 3, pp 527–545.
19. Rasim M.Alguliev, Ramiz M.Aliguliyev, and Nijat R.Isazade, Apr 2013, "Multiple documents summarization based on evolutionary optimization algorithm", Expert Systems with Applications, Elsevier, Volume 40, Issue 5, pp. 1675-1689.
20. Giang Tran, Eelco Herder, and Katja Markert, 2015, "Joint Graphical Models for Date Selection in Timeline Summarization", Proceedings of the 53rd Annual Meeting of the Association for Computational Linguistics and the 7th International Joint Conference on Natural Language Processing (Volume 1: Long Papers), Association for Computational Linguistics, pp. 1598–1607.
21. Libin Yang, Xiaoyan Cai, Yang Zhang, and Peng Shi, 2014, "Enhancing sentence-level clustering with ranking-based clustering framework for theme-based summarization", Information Sciences, Elsevier, Volume 260, pp. 37-50.
22. Melissa Ailem, Bowen Zhang, and Fei Sha, 2019, "Topic Augmented Generator for Abstractive Summarization", ArXiv.
23. Liu Na, Di Tang, Lu Ying, Tang Xiao-jun, and Wang Hai-wen, 2014, "Topic-Sensitive Multi-document Summarization Algorithm", Sixth International Symposium on Parallel Architectures, Algorithms and Programming, Beijing, pp. 69-74.
24. Lu Wang, and Claire Cardie, 2012, "Unsupervised Topic Modeling Approaches to Decision Summarization in Spoken Meetings", Proceedings of the 13th Annual Meeting of the Special Interest Group on Discourse and Dialogue, Association for Computational Linguistics, pp. 40-49.
25. Zongda Wu, Li Lei, Guiling Li, Enhong Chen, Hui Huang, Guandong Xu, and Chenren Zheng, 2017, "A Topic Modeling based Approach to Novel Document Automatic Summarization", Expert Systems with Applications: An International Journal, Pergamon Press, Inc., Volume 84, Number C, pp. 12-23.
26. Nikolaos Gialitsis, Nikiforos Pittaras, and Panagiotis Stamatopoulos, 2019, "A topic-based sentence representation for extractive text summarization", Proceedings of the Workshop MultiLing 2019: Summarization Across Languages, Genres and Sources, INCOMA Ltd., pp. 26–34.
27. Rada Mihalcea, and Paul Tarau, 2004, "TextRank: Bringing Order into Text", Proceedings of the 2004 Conference on Empirical Methods in Natural Language Processing, Association for Computational Linguistics, pp. 404–411.
28. Chin-Yew Lin, 2004, "ROUGE: A Package for Automatic Evaluation of Summaries", Text Summarization Branches Out, Association for Computational Linguistics, pp. 74–81.
29. https://www.wikihow.com/Main-Page
30. https://www.wikihow.com/Improve-the-Environment
31. Radim Rehurek, and Petr Sojka, 2010, "Software framework for topic modeling with large corpora", The LREC 2010 Workshop on New Challenges for NLP Frameworks, University of Malta, pp. 45-50.
32. McCallum, Andrew Kachites, 2002, "MALLET: A Machine Learning for Language Toolkit".
33. https://www.wikihow.com/Comb-Long-Hair